\documentclass[conference]{IEEEtran}
\IEEEoverridecommandlockouts    

\usepackage[paper=letterpaper,top=72pt,bottom=54pt,right=54pt,left=54pt]{geometry}
\usepackage{multirow}
\usepackage{lipsum}
\usepackage{amsmath}
\usepackage{amssymb}
\usepackage[ruled,vlined]{algorithm2e}
\usepackage{graphicx}
\graphicspath{{./Figures/}}

\usepackage{soul}
\usepackage{cite}
\usepackage{float}
\usepackage{caption}
\usepackage{subcaption}
\usepackage{stfloats}
\usepackage{amsmath,amssymb,amsfonts}
\usepackage{mathtools}
\usepackage{cite}
\usepackage{float}
\usepackage{caption}
\usepackage{subcaption}
\usepackage{stfloats}
\usepackage{amsmath,amssymb,amsfonts}
\usepackage{mathtools}
\usepackage{graphicx}
\usepackage{textcomp}
\usepackage{xcolor}
\usepackage{bm}
\usepackage{diagbox}
\usepackage{multirow}
\usepackage{makecell}
\usepackage{cases}
\usepackage[hidelinks]{hyperref}
\usepackage{lipsum}
\usepackage[british]{babel}
\usepackage[mathscr]{eucal}

\title{\LARGE \bf Physics-informed Teacher-student Ensemble Learning for Traffic State Estimation with a Varying Speed Limit Scenario}

\author{
	\parbox{\textwidth}{%
		\centering
		Archie J. Huang$^{1}$, Dongdong Wang$^{2}$, Shaurya Agarwal$^{3}$, Mohamed Abdel-Aty$^{3}$, Md Mahmudul Islam$^{3}$, Muhammad Shahbaz$^{3}$%
	}%
	\thanks{$^{1}$ Department of Building, Civil and Environmental Engineering, Concordia University, Montreal, Canada
		{\tt\small archie.huang@concordia.ca}}%
	\thanks{$^{2}$ Urban Artificial Intelligence Laboratory, University of Florida, Gainesville, USA
		{\tt\small dongdongwang@ufl.edu}}%
    \thanks{$^{3}$Department of Civil, Environmental and Construction Engineering, University of Central Florida, Orlando, USA
		{\tt\small \{Shaurya.Agarwal, M.Aty, Mahmudul, Muhammad.Shahbaz\}@ucf.edu}}%
}

\begin{document}
	
	\maketitle
	\thispagestyle{empty}
	\pagestyle{empty}
	
	\begin{abstract}
        Physics-informed deep learning (PIDL) neural networks have shown their capability as a useful instrument for transportation practitioners in utilizing the underlying relationship between the state variables for traffic state estimation (TSE). 
        Another efficient traffic management approach is implementing varying speed limits (VSLs) on transportation corridors to control traffic and mitigate congestion. 
        However, the existing training architecture of PIDL in the literature cannot accommodate the changing traffic characteristics on a freeway with VSL. 
        To tackle this challenge, we propose a novel framework integrating teacher-student ensemble training with PIDL neural networks for TSE under VSL scenarios. 
        The physics of flow conservation law is encoded locally in the teacher models by PIDL, and the student model uses a multi-layer perceptron classifier (MLP) to identify traffic characteristics and selects the ensemble member of PIDL neural networks for TSE. 
        %
        This integrated framework provides a natural solution for capturing the heterogeneity of VSL and accurately addressing the TSE problem.
        The case study results validate the proposed ensemble approach, demonstrating its superior performance in TSE compared to other popular baseline methods, as indicated by relative L2 error.

        \textbf{Index Terms —} Physics-informed Deep Learning, Ensemble Learning, Traffic State Estimation, Varying Speed Limit
	\end{abstract}

	\section{Motivation and Problem Formulation}
	\label{sec:Motivation and Problem Formulation}

    Traffic state estimation (TSE) is the process of estimating the current state of the traffic system, such as traffic flow $f$, speed $v$, and density $\rho$, based on observations of traffic conditions \cite{seo2017traffic}. It is a crucial process for effective traffic management and controls \cite{treiber2013traffic}. Physics-Informed Deep Learning (PIDL), also termed ``Physics-informed Neural Network (PINN)'', is a hybrid approach that combines the strengths of traditional physics-based models and machine learning techniques for solving partial differential equations (PDEs). In this approach, a neural network is trained to approximate the solution of a PDE subject to given boundary and initial conditions, while incorporating the underlying physical laws into the loss function used for training. 
    It has been demonstrated that PIDL can be adopted effectively in TSE to better perceive the traffic conditions with limited observation data \cite{huang2020physics, huang2022physics}.

Varying speed limits (VSLs) are a traffic management tool aimed at improving safety and reducing congestion on roads. Unlike fixed speed limits, which apply a single restriction to all drivers on a particular road, VSLs allow for real-time adjustments to speed limits based on various factors such as weather conditions, roadwork, and traffic flow. With VSL scenarios, the existing training architecture of PIDL cannot accommodate the changing physical equation governing the traffic flow. Therefore it is imperative to develop a new framework of PIDL neural networks for the purpose of TSE with the VSL scenario. The research objective of this work is listed as follows.

\textbf{Objective:} Develop a teacher-student ensemble of PIDL neural networks for traffic state estimation on a road infrastructure that has varying speed limits. The local traffic characteristics (i.e., speed limits) are stored in the locally deployed teacher neural networks, and the student model is a generalized PIDL neural network that can produce the estimation output of the entire road based on observation of the traffic.

\subsection{Problem Formulation}
\begin{figure}[htbp]
    \centerline{\includegraphics[width=0.5\textwidth]{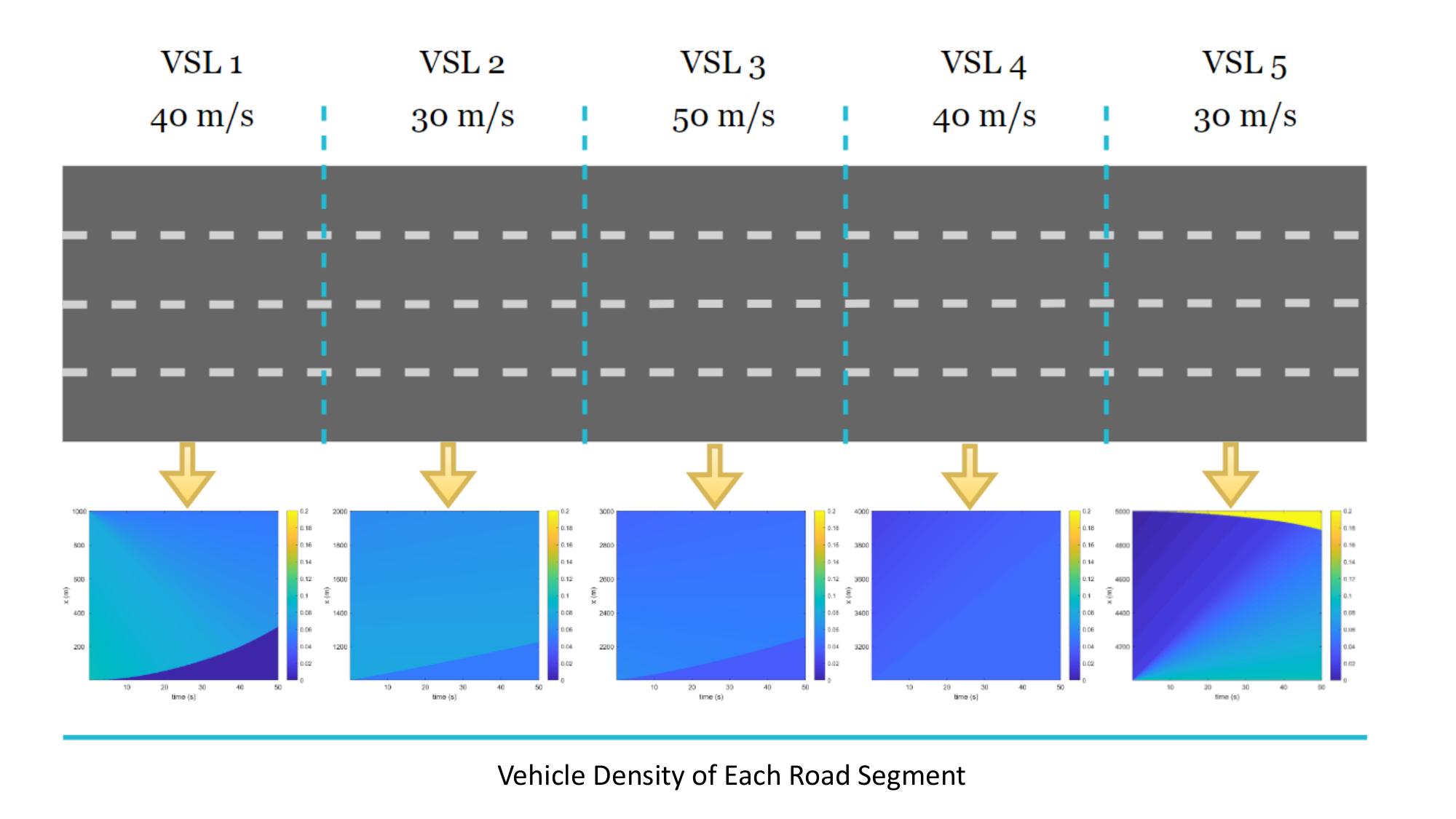}}
    \caption{Illustration of varying speed limit (VSL) scenario.}
    \label{fig:vsl_scenario}
\end{figure}

Let $\mathcal{D} = \mathcal{X} \times \mathcal{T}$ be a space-time domain and subdomain $\mathcal{D}^{\langle i \rangle} = \mathcal{X}^{\langle i \rangle} \times \mathcal{T}$ exists, in which $\mathcal{X}^{\langle i \rangle}$ are spatial segments of $\mathcal{X}$. For example, in Fig.~\ref{fig:vsl_scenario} the road $\mathcal{X}$ is split into five segments $\mathcal{X}^{\langle i \rangle}, i \in [1, 2, 3, 4, 5]$. The discretized homogeneous spatial units are $x^{\langle i \rangle [j]} \in \mathcal{X}^{\langle i \rangle}$. The temporal units are $t^{[k]} \in \mathcal{T}$. The value of the domain field is denoted as $\mathbf{V}(\mathcal{D})$, and $v(x, \, t)$ represents the value at $(x, \, t) \in \mathcal{D}$. 

On the segment $\mathcal{X}^{\langle i \rangle}$, scanty observations of the field value $v(x, \, t)$ at $(x_{o}^{\langle i \rangle [j]}, \, t_{o}^{[k]}) \in \mathcal{O}^{\langle i \rangle}$ are obtained. Collocation points are $(x_{c}^{\langle i \rangle [m]}, \, t_{c}^{[n]}) \in \mathcal{C}^{\langle i \rangle}$ where the values of the governing physical equation are computed. Collectively, $\mathcal{O}^{\langle i \rangle}$ in all segments constitute a domain $\mathcal{O} \subset \mathcal{D}$, and $\mathcal{C}^{\langle i \rangle}$ are assembled into $\mathcal{C} \subset \mathcal{D}$. The domain value of $\mathcal{O}$ is denoted as $\mathbf{V}(\mathcal{O})$, comprising the field observations $v(x_{o}^{\langle i \rangle [j]}, \, t_{o}^{[k]})$. The value of the governing physical equation at the collocation points is designated as $\mathbf{P}(\mathcal{C})$.

\textbf{Training of teacher PIDL neural networks:} On each segment $\mathcal{X}^{\langle i \rangle}$, given $\mathbf{V}(\mathcal{O}^{\langle i \rangle})$, the training of a teacher PIDL neural network becomes finding a mapping function $\mathsf{F_{tea}}(\cdot): \mathbf{V}(\mathcal{O}^{\langle i \rangle}) \rightarrow \mathbf{V_{T}}(\mathcal{O}^{\langle i \rangle})$, which minimizes the loss function of the teacher model $\mathsf{L_{tea}} = \mathsf{L_{est}}(\mathbf{V_{T}}(\mathcal{O}^{\langle i \rangle}), \mathbf{V}(\mathcal{O}^{\langle i \rangle})) + \mathsf{L_{phy}}(\mathbf{P}(\mathcal{C}^{\langle i \rangle}))$. The output of the traffic state from the teacher neural network is $\mathbf{V_{T}}(\mathcal{O}^{\langle i \rangle})$. Comparing this estimation output with the ground truth $\mathbf{V}(\mathcal{O}^{\langle i \rangle})$, we have the estimation-loss $\mathsf{L_{est}}$. The non-compliance of the governing physics is measured by the physics-loss $\mathsf{L_{phy}}$.

\textbf{Training of the student PIDL neural network:} The training of the student PIDL neural network is to find a mapping function $\mathsf{F_{stu}}(\cdot): \mathbf{V}(\mathcal{O}) \rightarrow \mathbf{V_{S}}(\mathcal{O})$, which minimizes the loss function of the student model $\mathsf{L_{stu}} = \mathsf{L_{est}}(\mathbf{V_{S}}(\mathcal{O}), \mathbf{V}(\mathcal{O}))$. In contrast to the teacher model which is trained with observation $\mathcal{O}^{\langle i \rangle}$ on the road segment $\mathcal{X}^{\langle i \rangle}$, the evaluation of the student model's estimation-loss $\mathsf{L_{est}}$ is with the entire training set $\mathcal{O} \subset \mathcal{X}$.

	\section{Background}
	\label{sec:Background}

    \subsection{Traffic State Estimation}

\begin{figure}[htbp]
    \centerline{\includegraphics[width=0.5\textwidth]{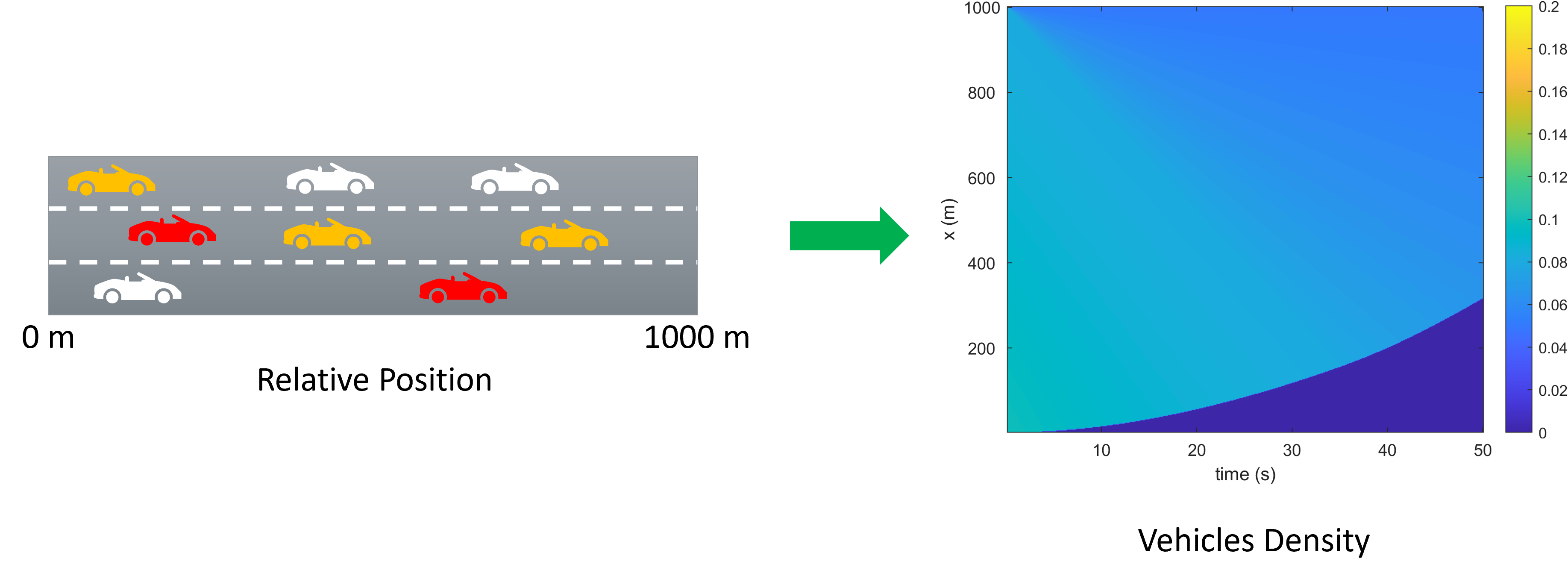}}
    \caption{From observations to spatiotemporal density.}
    \label{fig:traffic-state-estimation}
\end{figure}

Traffic state estimation (TSE) concerns the procedure of inferring traffic state variables based on limited data available, and the approaches for TSE can be categorized into model-based methods and data-based methods \cite{seo2016filter}. Model-based TSE relies on mathematical models to represent the underlying traffic behavior \cite{seo2016filter}. These models typically involve various parameters to simulate traffic behaviors. The advantages of model-based approaches include the ability to perform sensitivity analysis and to understand the underlying traffic mechanisms \cite{sumalee2011stochastic}. However, model-based approaches also have several disadvantages, including the difficulty in obtaining accurate model parameters \cite{tampere2007extended}, the need for regular model calibration \cite{wang2009adaptive}, and the potential for model errors and biases \cite{shahrbabaki2018data}. Data-based TSE, on the other hand, relies on direct observation of traffic data, obtained from sensors or other sources such as traffic cameras and GPS signals to estimate the traffic state \cite{leduc2008road}. It then applies machine learning algorithms to learn patterns and relationships in the data and predict the traffic state. Data-based TSE has the advantage of learning from historical data and adapting to changing traffic conditions. However, the need for large amounts of high-quality data, and handling missing and noisy data can be cumbersome and difficult for real-time applications \cite{antoniou2013dynamic}. Fig.~\ref{fig:traffic-state-estimation} depicts the process of estimating traffic density from observations of vehicle location on the road.

\subsection{Varying Speed Limit}

Varying speed limit (VSL) is implemented using dynamic signs that display the current speed limit, which can be adjusted by a central control system in response to changing conditions. VSL offers a flexible and effective way to manage traffic and improve road safety. One of the key advantages of VSL is the ability to respond to changing conditions in real-time \cite{zhang2017coordinated}. For example, during heavy rain or poor visibility, VSL can be used to lower speed limits to ensure drivers are traveling at a safe and appropriate speed. In addition, VSL can be adjusted to manage traffic during roadworks or lane closures, helping to reduce congestion and maintain a safe and efficient flow of traffic \cite{espitia2020event}. 

Potential drawbacks that are worth considering include the requirement of a sophisticated monitoring and control system to implement VSL, which can be expensive to install and maintain \cite{saidur2012applications}. Additionally, there is a risk of driver confusion or noncompliance if VSL is not clearly displayed or if drivers do not understand the reasoning behind the changes \cite{lee2006evaluation}. While there are challenges to implementation, VSL has the potential to make roads safer and more efficient for all road users and it has become an area of interest for engineers and researchers in the field of transportation \cite{karafyllis2017analysis}. This technology can provide significant benefits for road users, as it allows for more effective traffic management and reduces the likelihood of accidents and congestion.

\subsection{Physics-informed Deep Learning}

Deep learning methods are widely adopted on multiple frontiers of artificial intelligence and computational science, such as image perception \cite{lecun2015deep}, fluid simulation \cite{yang2016data}, and solving for a physical system \cite{geneva2019quantifying}. Employing such an approach to learn from and solve for a physical system involves approximating the parameterized analytical solution of the system \cite{lagaris1998artificial}.

Physics-Informed Deep Learning (PIDL) is a type of neural network that incorporates prior knowledge of physical laws and relationships into the optimization process. It allows for more accurate and efficient solutions since it utilizes the strength of the governing physical equations as a regularization agent in updating the weight and bias parameters of the neural network \cite{raissi2017physics}. PIDL can solve complex and nonlinear PDEs that are difficult for traditional numerical methods or analytical solutions \cite{raissi2017physics2}, or trained with limited or noisy data \cite{haghighat2021physics}. 



\subsection{Ensemble Learning}

Ensemble learning (EL) is a machine learning technique that combines the predictions of multiple models to obtain a more accurate and robust result \cite{dong2020survey}. Combining the results of different models, EL limits the over-fitting issue from individual models in the final output, improves the generalization ability of the model, and mitigates the impact of noisy data \cite{huang2009research}. Its various applications include object detection \cite{wu2012object}, activity recognition \cite{hamid2012classifier}, and remote sensing \cite{polikar2012ensemble}.

Each of the base models in EL can be trained by using various algorithms or on different subsets of the training data. One common ensemble learning approach for traffic state estimation is the use of a combination of model-based and data-driven methods. For example, a model-based approach may be used to incorporate prior knowledge of physical laws and relationships, such as the fundamental diagram of traffic flow, while a data-driven approach may be used to handle the complexity and variability of real-world traffic data. The combination of these two approaches can provide a more robust and accurate solution compared to either approach alone. Another approach is to use an ensemble of different neural network architectures, such as feedforward networks, recurrent networks, and convolutional networks. Each network can be trained on a different subset of the observation input or using different training algorithms, and the final prediction can be obtained by combining the predictions of each network.

\section{Methodology - Teacher-Student Ensemble of PIDL for TSE}

Traffic state estimation involves the prediction of traffic variables, such as vehicle density, speed, and flow, based on traffic observations. Due to the installation and maintenance costs of embedded sensors and cameras along the road infrastructures, the main challenge is that the available data for training a TSE model is often noisy and limited. PIDL incorporates prior knowledge of physical laws and relationships, such as the fundamental diagram (FD) of traffic flow, into the training process of the neural network. The governing physical equations from the traffic flow theory describe the relationship between traffic state variables (e.g., vehicle density and speed). This allows the neural network optimization to be guided by the underlying physical laws, rather than relying solely on data-driven methods. The fundamentals of traffic flow theory is given below.

\subsection{Physics of Traffic Flow}

Given its simplicity, we select the Greenshields fundamental diagram (FD) \cite{greenshields1935study} to depict the relationship between traffic state variables speed $v$ and vehicle density $\rho$, given by \eqref{eqn:greenshields}, in which the maximum density is $\rho_m$, and the free-flow speed is $v_f$.

\begin{equation} \label{eqn:greenshields}
    v(x, \, t)  =  v_f - \frac{v_f}{\rho_m} \cdot \rho(x, \, t)
\end{equation}

Lighthill-Whitham-Richards (LWR) conservation law \cite{lighthill1955kinematic} is chosen as the physical model dictating vehicular traffic and is formulated as \eqref{eqn:lighthill}. The traffic flow at location $x$ and timestamp $t$ is $q(x, \, t)$. Since FD relates the traffic state variables such as density and velocity, pairing LWR with Greenshields FD in \eqref{eqn:greenshields} can express the flow conservation law as \eqref{eqn:lwr_density}, with density $\rho(x, \, t)$ as the sole independent variable in the partial differential equation (PDE).

\begin{equation} \label{eqn:lighthill}
     \frac{\partial \rho(x, \, t)}{\partial t} + \frac{\partial q(x, \, t)}{\partial x} = 0
\end{equation}

\begin{equation} \label{eqn:lwr_density}
    \frac{\partial \rho(x, \, t)}{\partial t} + v_f \left(1 - \frac{2\rho(x, \, t)}{\rho_m}\right)  \frac{\partial \rho(x, \, t)}{\partial x} = 0
\end{equation}

\subsection{Training a PIDL Neural Network}

The cost function of a PIDL neural network encodes the physics of traffic flow in the physics cost term $J_{PHY}$. Together with the deep learning cost $J_{DL}$, which measures the accuracy of the estimation output, these two cost terms constitute the cost function of PIDL, which uses the parameter $\mu$ to adjust the weights of $J_{DL}$ and $J_{PHY}$ as in \eqref{eqn:pidl_cost}.

\begin{equation} \label{eqn:pidl_cost}
    J = \mu * J_{PHY} + (1 - \mu) * J_{DL}
\end{equation}

The \textbf{physics cost term $J_{PHY}$} is computed on the collocation points $(x_{c}, t_{c}) \in \mathrm{C}$, taking the physics of flow conservation expressed by the density variable in \eqref{eqn:lwr_density}, the physics cost measured by mean squared error (MSE) is formulated in \eqref{eqn:physics_cost}, in which the estimation of density $\rho(x_{o}^{j}, \, t_{o}^{j})$ at the observation point $(x_{o}^{j}, \, t_{o}^{j})$ is denoted as $\hat{\rho}(x_{o}^{j}, \, t_{o}^{j})$.

\begin{equation} \label{eqn:physics_cost}
\begin{split}
        J_{PHY} = \frac{1}{N_{c}}\sum_{i=1}^{N_{c}} & \bigg| \frac{\partial \hat{\rho}(x_{c}^{i}, \, t_{c}^{i})}{\partial t} \\
        & + v_f\left(1 - \frac{2 \hat{\rho}(x_{c}^{i}, \, t_{c}^{i})}{\rho_m}\right)  \frac{\partial \hat{\rho}(x_{c}^{i}, \, t_{c}^{i})}{\partial x}\bigg|^2
\end{split}
\end{equation}

Similarly, the \textbf{deep learning cost $J_{DL}$} by MSE at observation points $(x_{o}, t_{o}) \in \mathrm{O}$ is shown in \eqref{eqn:deep_learning_cost}.

\begin{equation} \label{eqn:deep_learning_cost}
    J_{DL} = \frac{1}{N_{o}}\sum_{j=1}^{N_{o}}\left|\rho(x_{o}^{j}, \, t_{o}^{j}) - \hat{\rho}(x_{o}^{j}, \, t_{o}^{j})\right|^2  
\end{equation}

For additional technical details about training a PIDL neural network for traffic state estimation, we direct readers' attention to the literature \cite{huang2022physics, shi2022physics, huang2023limitations, di2023physics}.

\subsection{Teacher-student Ensemble}

\begin{figure}[htbp]
    \centerline{\includegraphics[width=0.46\textwidth]{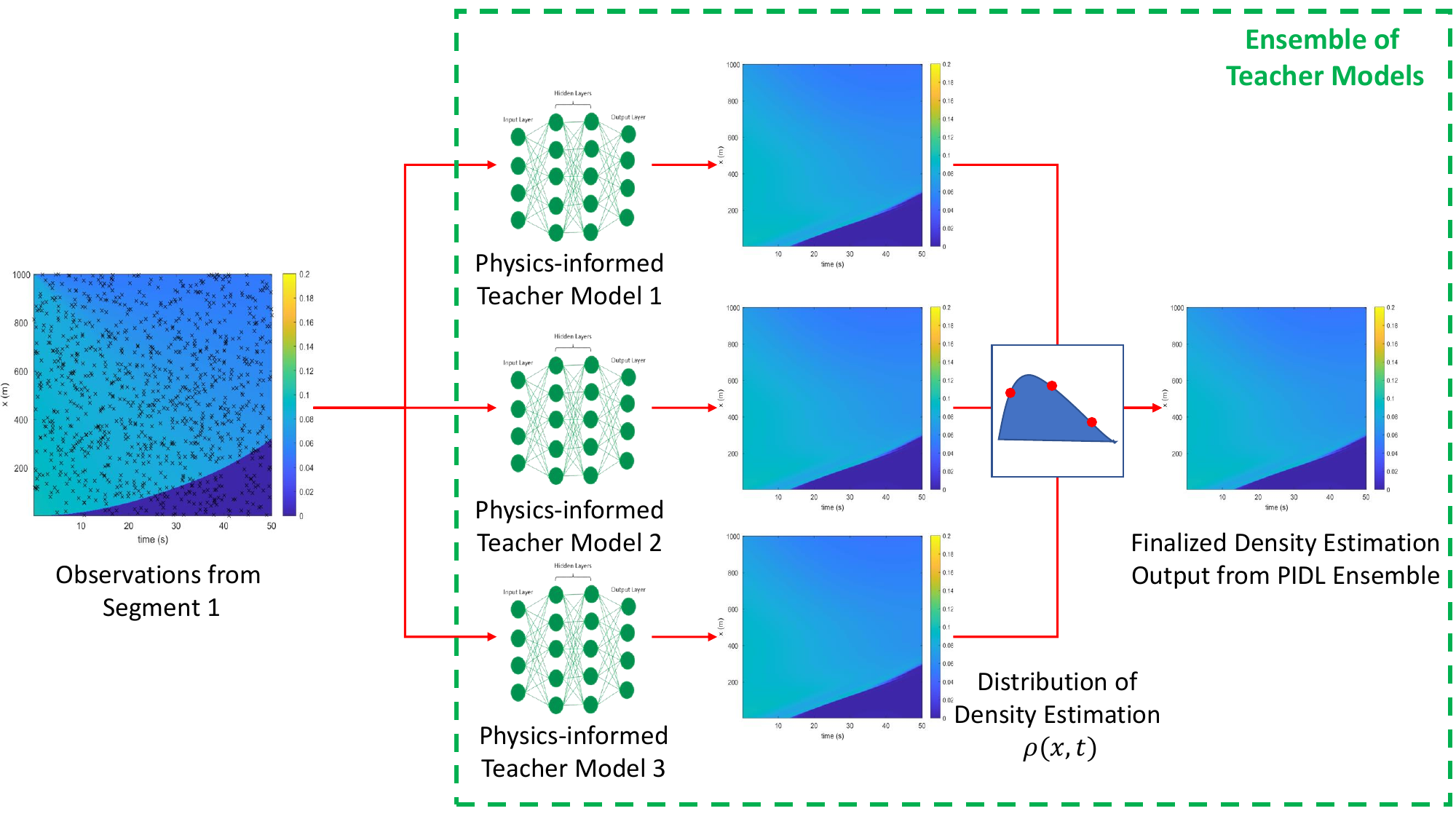}}
    \caption{Training an ensemble of physics-informed teacher models.}
    \label{fig:training_teacher_model}
\end{figure}

The training of a teacher PIDL neural network involves using the observation from a local road segment in which the traffic characteristic (e.g., free-flow speed $v_f$) is consistent over a given time. The local traffic characteristics are encoded within the teacher PIDL neural network, and the task of \textbf{training an ensemble of teacher models} is to sample a plethora of physics-informed teacher neural networks with a range of initialization configurations. Fig.~\ref{fig:training_teacher_model} illustrates the training process in which the finalized output from teacher models takes the weighted average of the density estimation from the ensemble. 

\begin{figure}[htbp]
    \centerline{\includegraphics[width=0.48\textwidth]{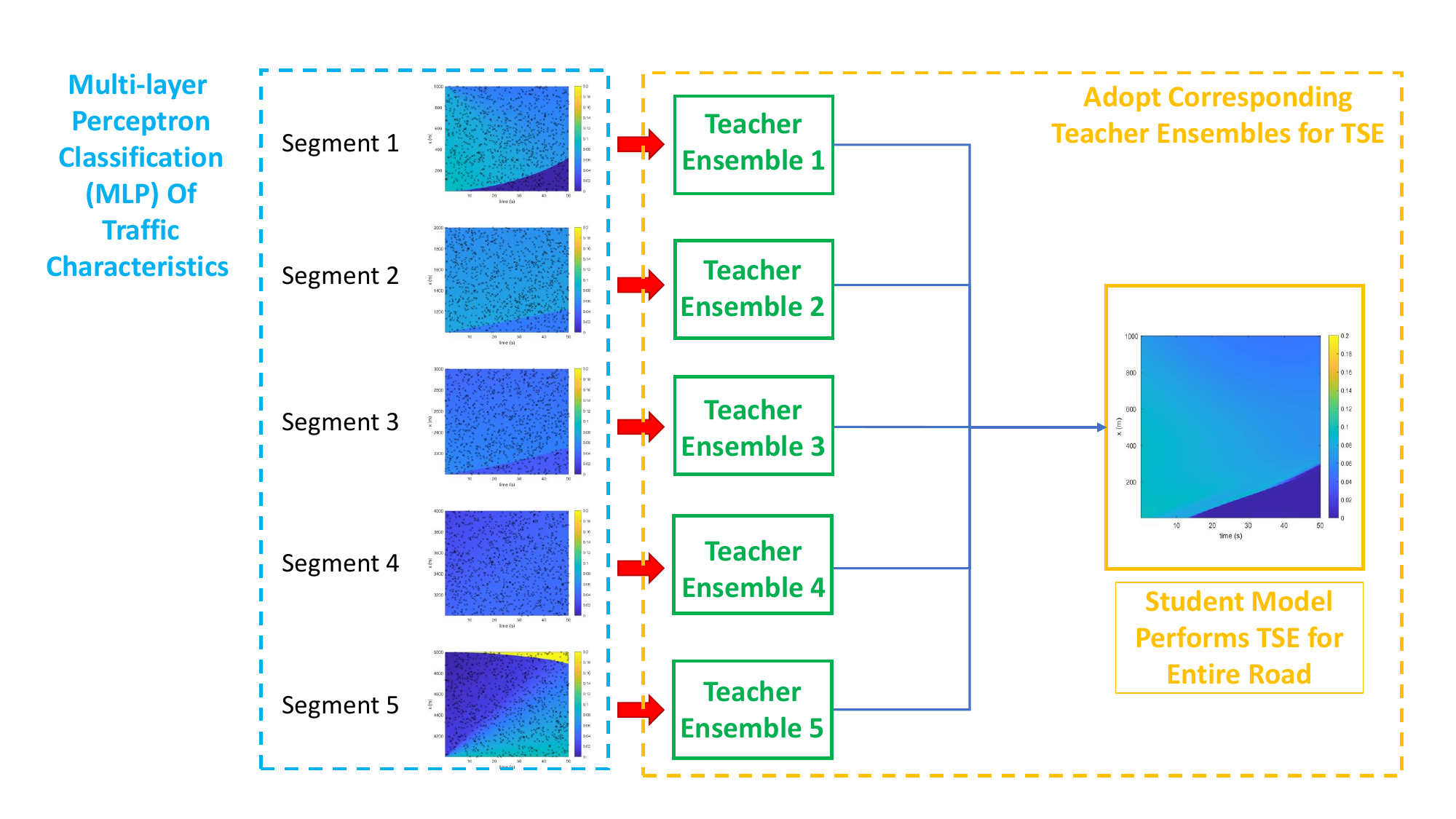}}
    \caption{Two-step approach for training the student model.}
    \label{fig:training_student_model}
\end{figure}

The \textbf{training of the student model} involves a two-step approach: the \textit{first step} is to adopt a multi-layer perceptron classifier (MLP) to extract the local traffic characteristics of each segment based on scant traffic observations. The \textit{second step} is to build a generalized model to adopt the corresponding ensemble of teacher PIDL neural networks based on the extracted traffic information from MLP.  This two-step approach enables the student model to perform the TSE task for the entire road with varying speed limits, and the process is demonstrated in  Fig.~\ref{fig:training_student_model}.

The loss function of the student PIDL neural network $\mathsf{L_{stu}}$ is represented by the estimation-loss $\mathsf{L_{est}}(\mathbf{V_{S}}(\mathcal{O})$. In the following case study, this estimation-loss is measured by the relative percent $\mathrm{L}_2$ error formulated in \eqref{eqn:l2_error}, with $\hat{\rho}(x^{k}, t^{k})$ representing the estimation of density at $(x^{k}, t^{k})$.

\begin{equation} \label{eqn:l2_error}
    \mathrm{L}_2^{error}
    = \frac{\sqrt{\sum_{k=1}^{N_1 \cdot N_2} \left|\hat{\rho}(x^{k}, t^{k}) - \rho(x^{k}, t^{k})\right|^2}}{\sqrt{\sum_{k=1}^{N_1 \cdot N_2}\left|\rho(x^{k}, t^{k})\right|^2}} 
\end{equation}
	
	\section{Case Study}
	\label{sec:Case Study}

	\subsection{Experimental Design}

A $5000$-meter test-bed is designed to simulate traffic flow, and the traffic density is solved by using the Lax-Hopf method \cite{mazare2011analytical}. The test-bed contains 5 sequential $1000$-meter road segments with varying speed limits. The simulation runs for $50$ seconds, under the traffic characteristics (speed limit $v_f$, jam density $\rho_{m}$, upstream flow $f_{up}$, and downstream flow $f_{down}$) of each road segment listed in Table~\ref{tab:traffic_characteristics}. The first road segment is from $x = 0$ to $x = 1000$. The unit of free-flow speed value $v_f$ is meter per second, and the unit of maximum density value $\rho_m$ is vehicle per meter. The units of the upstream flow $f_{up}$ and downstream flow $f_{down}$ are both vehicles per second. 

\begingroup
\renewcommand{\arraystretch}{1.2} 
    \begin{table}[htbp]
        \caption{Traffic Characteristics of Each Road Segment}
        \begin{center}
            \begin{tabular}{|p{1cm}|p{1cm}|p{1cm}|p{1cm}|p{1cm}|}\hline
                \multirow{2}{*}{\thead{\textbf{Road} \\ \textbf{Segment}}} &  \multicolumn{4}{c|}{\textbf{Value of Parameters}} \\
                \cline{2-5}
                & $v_f$ & $\rho_m$  & $f_{up}$ & $f_{down}$ \\\hline
                \textbf{1}  &  40  & 0.10 &  0  & 1.0 \\\hline
                \textbf{2}  &  30  & 0.15 &  1.0  & 1.125 \\\hline
                \textbf{3}  &  50  & 0.10 &  1.125  & 1.25 \\\hline
                \textbf{4}  &  40  & 0.15 &  1.25  & 1.5 \\\hline
                \textbf{5}  &  30  & 0.10 &  1.5  & 0 \\\hline
            \end{tabular}
            \label{tab:traffic_characteristics}
        \end{center}
    \end{table}
\endgroup

\begin{figure}[htbp]
    \centerline{\includegraphics[width=0.4\textwidth]{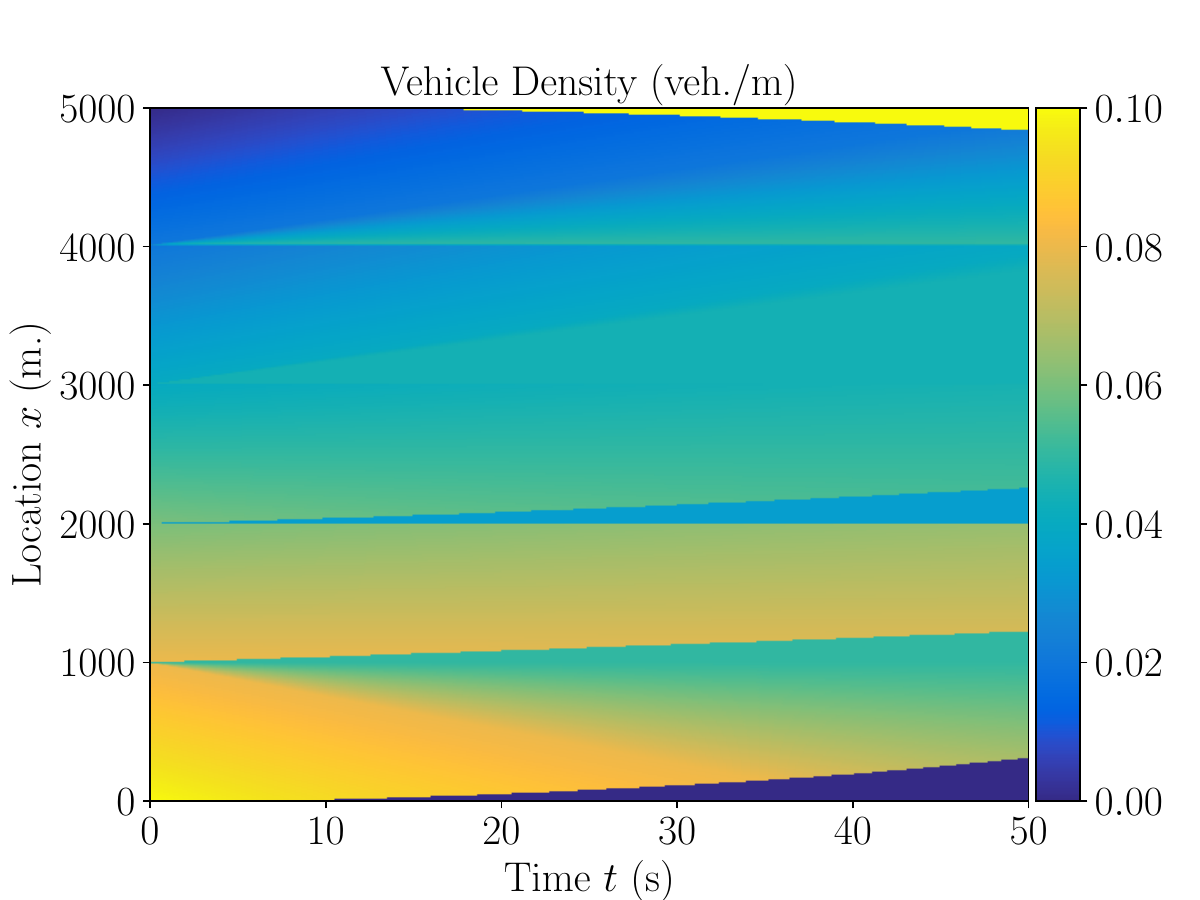}}
    \caption{Illustration of ground-truth vehicle density values. }
    \label{fig:density_data}
\end{figure}

The density values in this dataset are illustrated in Fig.~\ref{fig:density_data}. The spatial resolution of the density dataset $\Delta x$ is $2$-meter and the temporal resolution of the dataset $\Delta t$ is $0.1$-second.  

\subsection{Training the PIDL Ensemble and the Benchmark Methods}

The density dataset illustrated in Fig.~\ref{fig:density_data} contains over a million data points (1,253,001). In each road segment, we select 50 data points to train the multi-layer perceptron classifier (MLP) to extract traffic characteristics and use another 250 density values as traffic observations for the student model to reconstruct the density field. Together, these 1500 ($(50 + 250) \times 5$) data points represent a mere $0.12\%$ of the dataset for training the PIDL teacher-student ensemble.

Additionally, we select (1) the non-ensemble PIDL neural network, (2) the physics-uninformed Deep Learning (DL) neural network, and (3) the long short-term memory with interpolation (LSTM + interp2) as the benchmark methods for comparison on TSE results. The \textit{non-ensemble PIDL} does not include the ensemble learning process in Fig.~\ref{fig:training_teacher_model}, instead relying on a single PIDL model for local traffic state estimation in each road segment. The \textit{DL neural network} is a regular fully connected neural network with the same architecture as PIDL, without the knowledge of traffic flow dynamics. Lastly, the \textit{long short-term memory with interpolation} utilizes sequential observations of the traffic state for TSE.

\subsection{Results and Discussion}

\begin{figure*}[htbp]
     \centering
     \begin{subfigure}[b]{0.42\textwidth}
         \centering
         \includegraphics[width=\textwidth]{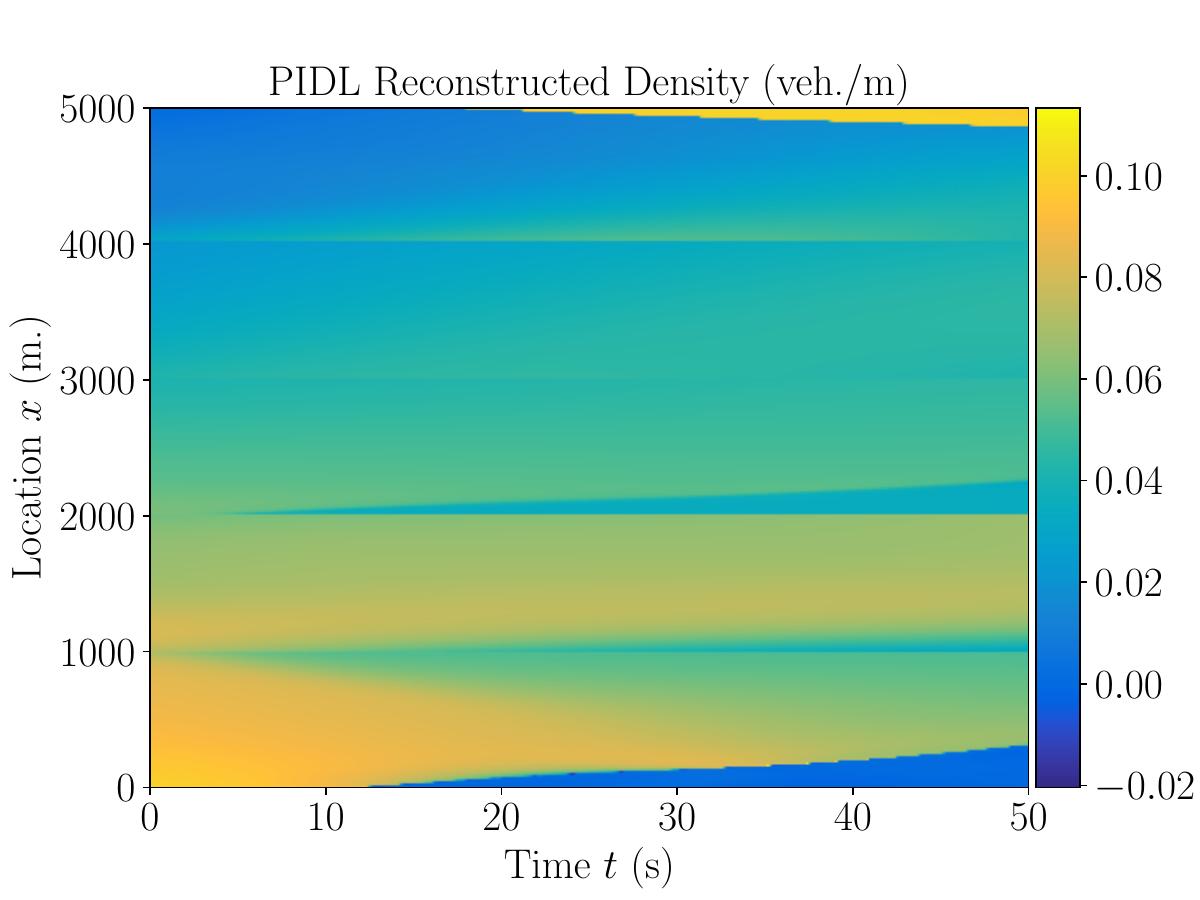}
    \caption{PIDL Teacher-Student Ensemble}
    \label{fig:vehicle_density_ensemble}
     \end{subfigure}
     \hfill
     \begin{subfigure}[b]{0.42\textwidth}
         \centering
         \includegraphics[width=\textwidth]{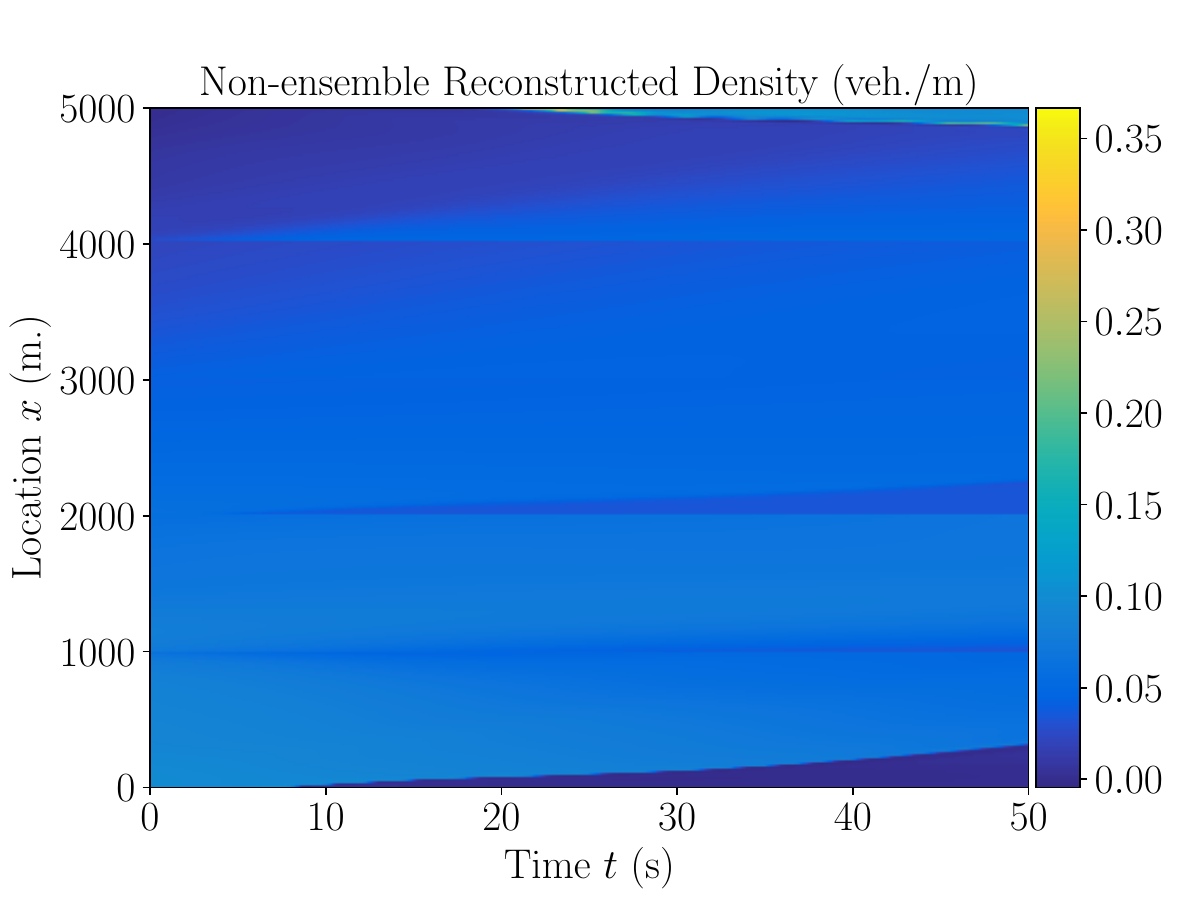}
    \caption{Non-ensemble PIDL}
    \label{fig:vehicle_density_dl_ensemble}
     \end{subfigure}
    \hfill
     \begin{subfigure}[b]{0.42\textwidth}
         \centering
         \includegraphics[width=\textwidth]{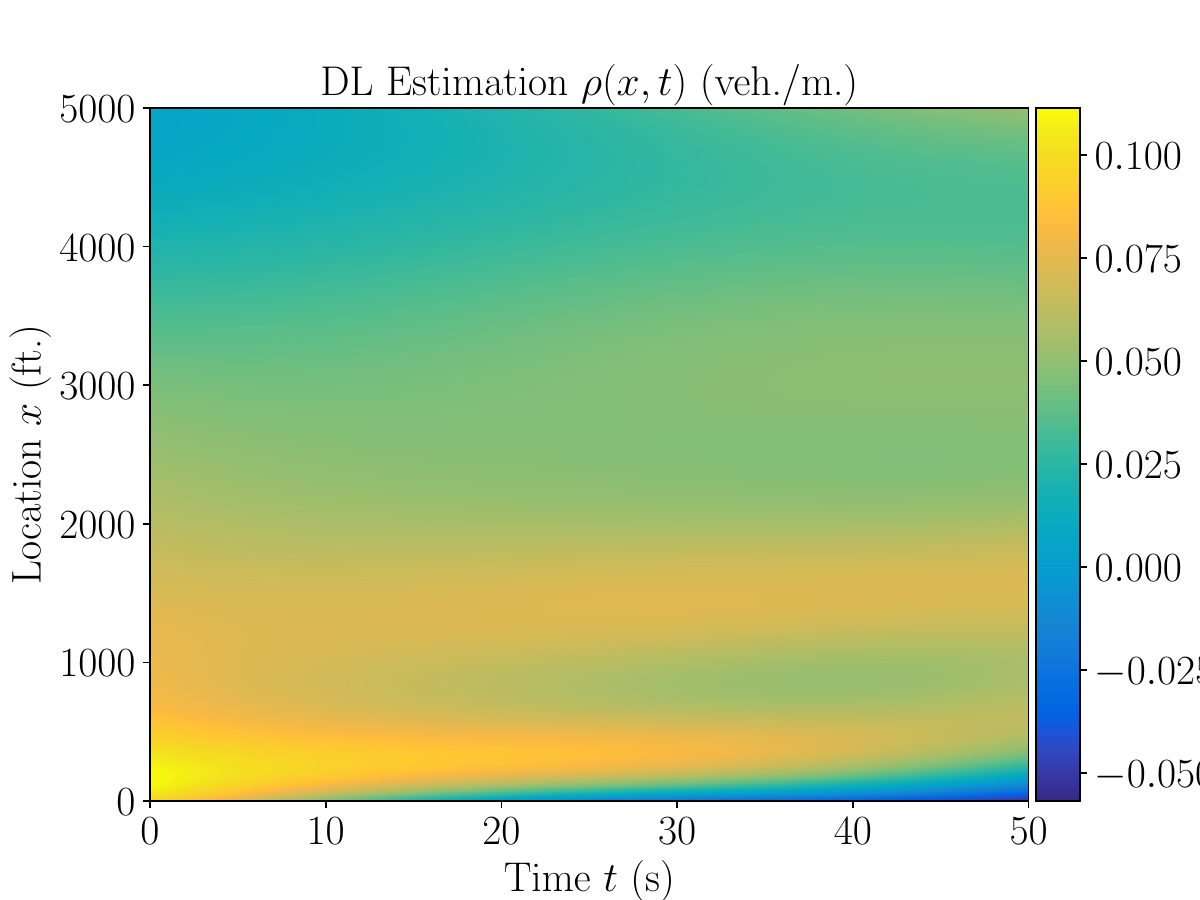}
    \caption{DL Neural Network}
    \label{fig:vehicle_density_ensemble}
     \end{subfigure}
     \hfill
     \begin{subfigure}[b]{0.42\textwidth}
         \centering
         \includegraphics[width=\textwidth]{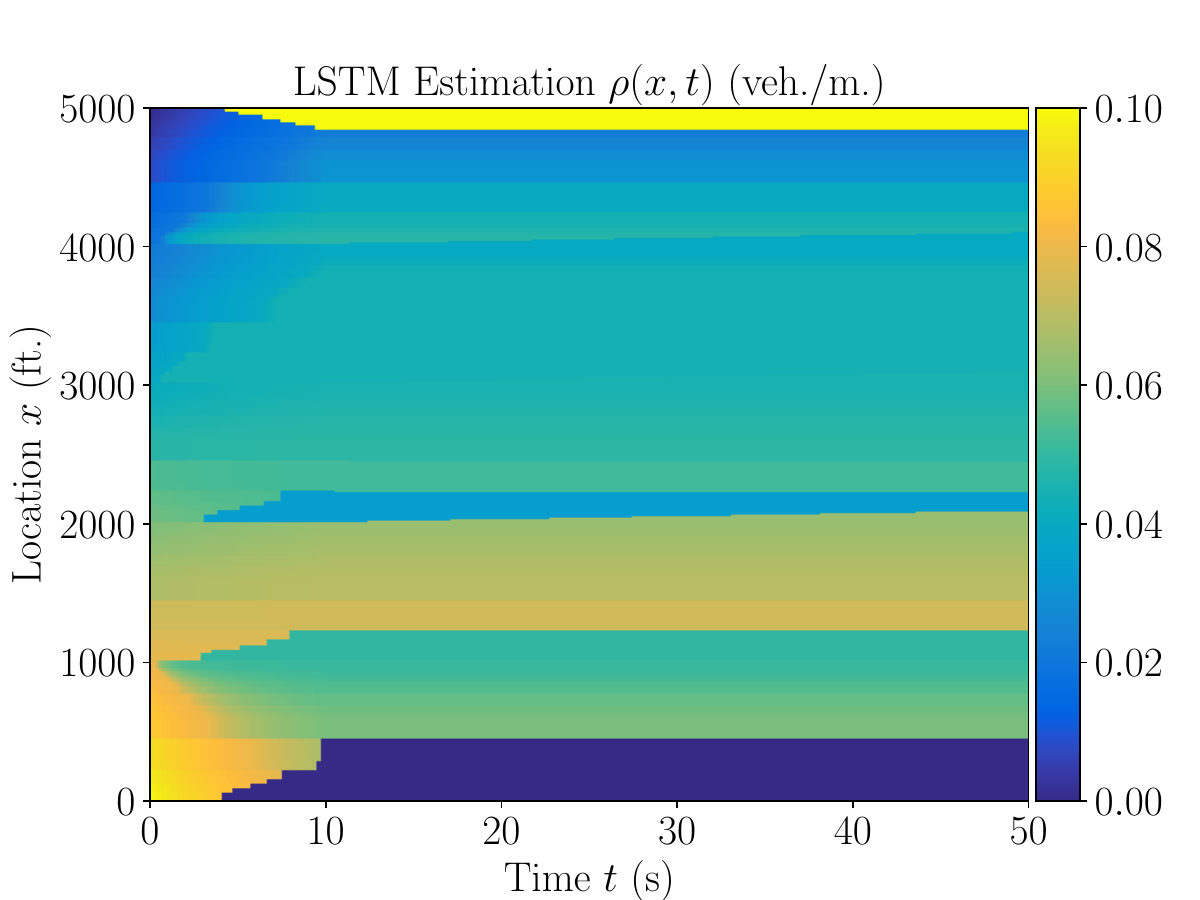}
    \caption{LSTM with Interpolation}
    \label{fig:vehicle_density_dl_ensemble}
     \end{subfigure}
    \caption{Estimation Results of Test-bed Traffic Density $\rho(x, \, t)$}
    \label{fig:vehile_density_tse_result}
\end{figure*}

The TSE results of the traffic density on the test-bed by the (1) PIDL teacher-student ensemble, (2) non-ensemble PIDL, (3) DL neural network, and (4) LSTM with interpolation are tabulated in Table~\ref{tab:vehile_density_tse_result}. The TSE accuracy is measured by the relative percent $\mathcal{L}_2$ error defined in \eqref{eqn:l2_error}. The reconstruction of the density field by the above four methods are also shown in Fig.~\ref{fig:vehile_density_tse_result}. 

From Table~\ref{tab:vehile_density_tse_result}, our proposed approach of PIDL teacher-student ensemble outperforms all benchmark models, achieving the lowest relative percent $\mathcal{L}_2$ error of $0.0389$. The non-ensemble version of the PIDL neural network is the best model among the benchmark methods, with the relative percent $\mathcal{L}_2$ error at $0.0624$. The LSTM is the worst benchmark method in this case, with a staggering relative percent $\mathcal{L}_2$ error of $0.408$. 

\begingroup
\renewcommand{\arraystretch}{1.2} 
    \begin{table}[htbp]
        \caption{Relative $\mathcal{L}_2$ Error of TSE Results on Test-bed Traffic Density}
        \begin{center} 
            \begin{tabular}{|p{0.6cm}|p{4cm}|p{2.4cm}|}\hline
                & \textbf{Model Type} & \textbf{Relative $\mathcal{L}_2$ Error}\\\hline
                \textbf{1}  &  PIDL Teacher-Student Ensemble  & \textbf{3.89e-02} \\\hline
                \textbf{2}  &  Non-ensemble PIDL  &  6.24e-02 \\\hline
                \textbf{3}  &  Deep Learning   &  1.47e-01 \\\hline
                \textbf{4}  &  LSTM with Interpolation & 4.08e-01  \\\hline
            \end{tabular}
            \label{tab:vehile_density_tse_result}
        \end{center}
    \end{table}
\endgroup

Observing the Fig.~\ref{fig:vehile_density_tse_result}, both (1) PIDL teacher-student ensemble, and (2) non-ensemble PIDL capture the unique traffic patterns of each road segment. The (3) DL neural network without the knowledge of physics cannot distinguish the varying traffic characteristics in each segment, underfitting the density dataset, and the (4) LSTM method with interpolation does not satisfactorily capture the traffic patterns in the reconstruction.

 \section{Conclusion}

This work proposes a physics-informed teacher--student ensemble learning framework for traffic state estimation (TSE) under varying speed limit (VSL) scenarios. By combining locally specialized teacher PIDL models with a student model that classifies traffic characteristics and selects the appropriate ensemble member, the proposed architecture overcomes the limitations of conventional PIDL approaches that assume spatially homogeneous traffic dynamics.

The case study on a 5000-meter synthetic test-bed with five sequential VSL segments demonstrates the effectiveness of the framework. Using only 0.12\% of the full dataset for training, the proposed teacher--student ensemble achieved a relative percent $L_2$ error of $3.89\times10^{-2}$, outperforming the non-ensemble PIDL ($6.24\times10^{-2}$), the physics-uninformed deep learning model ($1.47\times10^{-1}$), and the LSTM with interpolation ($4.08\times10^{-1}$). These results confirm that integrating physics guidance with ensemble specialization substantially improves estimation accuracy in heterogeneous traffic environments.

Despite its strong performance, several limitations remain. The current framework assumes piecewise-constant traffic characteristics within predefined road segments and relies on the Greenshields fundamental diagram, which may not fully capture complex or non-equilibrium traffic behavior. Furthermore, the validation is conducted on synthetic LWR-generated data; performance under real-world noisy measurements and dynamically changing VSL policies requires further investigation. 

Future work will focus on extending the framework to field data, enhancing the adaptability of the student model to continuously evolving traffic conditions, and developing online updating strategies for real-time intelligent transportation system deployment.
    
	\bibliographystyle{IEEEtran}
	\bibliography{ensemble} 

\end{document}